\documentclass{article}

\usepackage{PRIMEarxiv}

\usepackage[utf8]{inputenc} 
\usepackage[T1]{fontenc}    
\usepackage{hyperref}       
\usepackage{url}            
\usepackage{booktabs}       
\usepackage{amsfonts}       
\usepackage{nicefrac}       
\usepackage{microtype}      
\usepackage{lipsum}
\usepackage{fancyhdr}       
\usepackage{graphicx}       
\graphicspath{{media/}}     

\usepackage{booktabs} 

\usepackage{graphicx}
\usepackage{array}
\usepackage{times}
\usepackage{epsfig}
\usepackage{amsmath}
\usepackage{outlines}
\usepackage{soul}
\usepackage{hyperref}
\usepackage{natbib}
\usepackage{url}
\usepackage{makecell}
\usepackage{amssymb}
\usepackage{xparse}
\usepackage{float}
\usepackage{subcaption}
\usepackage{algpseudocode}
\usepackage{changepage}
\usepackage{makecell}

\usepackage[table,x11names,usenames,dvipsnames]{xcolor}
\newcolumntype{2}{!{\vrule width 1.5pt}}

\definecolor{LightGray}{gray}{0.9}
\definecolor{MediumGray}{gray}{0.72}
\definecolor{LightGreen}{rgb}{0.56, 0.93, 0.56}
\definecolor{PastelRed}{rgb}{1.0, 0.41, 0.38}
\definecolor{PastelOrange}{rgb}{1.0, 0.7, 0.28}

\NewDocumentCommand{\rot}{O{45} O{1em} m}{\makebox[#2][l]{\rotatebox{#1}{#3}}}%

\title{The Power of Next-Frame Prediction for Learning Physical Laws
\thanks{\textit{\underline{Intended for publication}}: 
} 
}

\author{
  Thomas Winterbottom\textsuperscript{1}\\
  Department of Computer Sciences\\
  Durham University \\
  Durham\\
  \texttt{thomas.i.winterbottom@durham.ac.uk} \\
    \And
  G. Thomas Hudson\textsuperscript{1}\\
  Department of Computer Sciences\\
  Durham University \\
  Durham\\
  \texttt{g.t.hudson@durham.ac.uk} \\
    \And
  Daniel Kluvanec\textsuperscript{1}\\
  Department of Computer Sciences\\
  Durham University \\
  Durham\\
  \texttt{daniel.kluvanec@durham.ac.uk} \\
    \And
  Dean Slack\\
  Department of Computer Sciences\\
  Durham University \\
  Durham\\
  \texttt{dean.l.slack@durham.ac.uk} \\
    \And
  Jamie Sterling\\
  Department of Computer Sciences\\
  Durham University \\
  Durham\\
  \texttt{jamie.sterling@durham.ac.uk} \\
    \And
  Junjie Shentu\\
  Department of Computer Sciences\\
  Durham University \\
  Durham\\
  \texttt{junjie.shentu@durham.ac.uk} \\
    \And
  Chenghao Xiao\\
  Department of Computer Sciences\\
  Durham University \\
  Durham\\
  \texttt{junjie.shentu@durham.ac.uk} \\
    \And
  Zheming Zhou \\
  School of Computing\\
  Newcastle University\\
  Newcastle\\
  \texttt{zheming.zuo@newcastle.ac.uk} \\
    \And
  Noura Al Moubayed \\
  Department of Computer Sciences\\
  Durham University \\
  Durham\\
  \texttt{noura.al-moubayed@durham.ac.uk} \\
}

\begin{document}
\maketitle

\begin{abstract}
Next-frame prediction is a useful and powerful method for modelling and understanding the dynamics of video data. Inspired by the empirical success of causal language modelling and next-token prediction in language modelling, we explore the extent to which next-frame prediction serves as a strong foundational learning strategy (analogous to language modelling) for inducing an understanding of the visual world. In order to quantify the specific visual understanding induced by next-frame prediction, we introduce six diagnostic simulation video datasets derived from fundamental physical laws created by varying physical constants such as gravity and mass. We demonstrate that our models trained \textit{only} on next-frame prediction are capable of predicting the value of these physical constants (\textit{e.g.} gravity) \textit{without} having been trained directly to learn these constants via a regression task. We find that the generative training phase \textit{alone} induces a model state that can predict physical constants significantly better than that of a random model, improving the loss by a factor of between 1.28 to 6.24.
We conclude that next-frame prediction shows great promise as a general learning strategy to induce understanding of the many `laws' that govern the visual domain \textit{without} the need for explicit labelling.
\end{abstract}


\section{Introduction}
\begin{figure}[!h]
	\centering
	\includegraphics[width=0.6\textwidth]{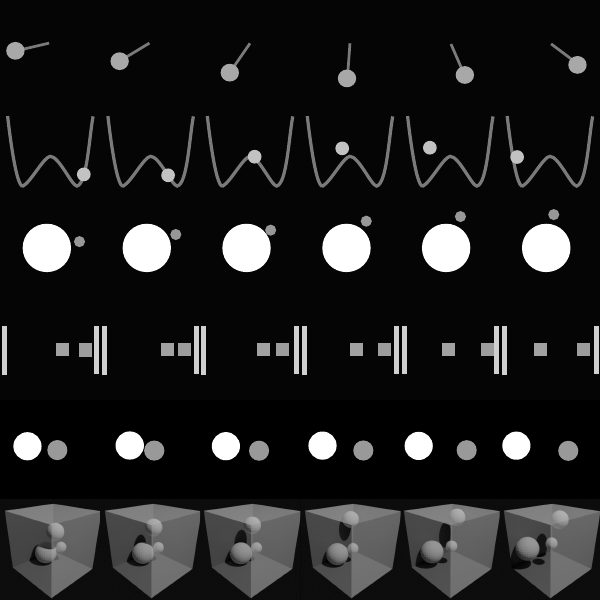}
	\caption{Example visualisations of 6 sequential frames from each of our six proposed dynamic simulation video datasets. From top to bottom: \textbf{Pendulum}, \textbf{Roller} coaster with flight, Mars \textbf{Moon}, Colliding \textbf{Blocks}, \textbf{2D Bouncing} balls, and \textbf{3D bouncing} balls.}
	\label{fig_dataset_overview}
\end{figure}

Generative video prediction is a popular and active area of research \citep{Castrejon_2019_ICCV,Oprea2020ARO,9063513} that has recently adopted transformer-based architectures \citep{carion2020end,Rakhimov2021LatentVT,Farazi2019FrequencyDT,farazi2021local} that have powered recent innovations in language modelling. The empirical success of word token prediction as a pretraining task is encouraging for the collective goal of ``general understanding'' in machine learning. If a model can learn to always predict the correct missing text token, or always accurately generate the next-frame in a video, can we not be satisfied that it understands text and vision well enough to model the `laws' governing the real world? Further still, such a training regime could circumvent the need to train a model to explicitly learn the underlying laws exhibited in the data. We don't need to create individual datasets for each `law' we wish to teach the model (\textit{e.g.} gravity encoded in vision or the concept of sarcasm in language) if success at generative pretraining will elegantly \textit{require} an understanding of \textit{all} such laws. This is evident and well-studied in the linguistic domain, where language models have shown to generalise to an impressive range of downstream tasks, and at scale exhibit emergent properties beyond their pretraining task \cite{wei2022emergent, brown2020language}. But can \textit{visual} models infer a `real understanding' of the underlying laws of its input data modality \textit{without} explicitly being taught them? If self-supervised pretraining can implicitly learn an approximation of these laws, can we easily extract this information to quantify any learned understanding? In this paper, we aim for a careful and diagnostic approach to answer these questions. To this end, we introduce six dynamic simulation datasets for video prediction pretraining (see Figure \ref{fig_dataset_overview}) each paired with at least one `probing' task. 

These probing tasks allow us to quantifiably measure the model's internal understanding of a physical law in the video sequence, \textit{e.g.} a loss on the task of predicting the gravity observed in a video clip. 

We evaluate these datasets on both fully convolutional 2D \textit{`CNN'} and a \textit{`Patch Transformer'} using convolutions and transformer blocks. Our probing experiments demonstrate that self-supervised pretraining on next-frame prediction significantly improves performance on our simulation understanding tasks. These results on our diagnostic datasets show the potential of the inductive power of generative pretraining for vision. Our implementation is available on GitHub\footnote{\url{https://github.com/Visual-modelling}}
\begin{figure*}
    \centering
    \includegraphics[width=1.0\textwidth]{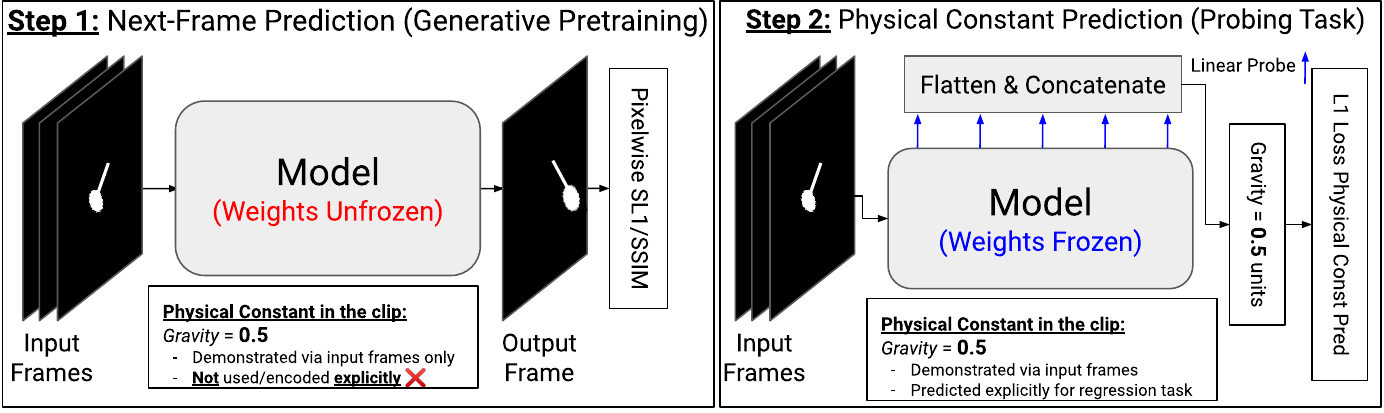}
    \caption{The two steps of training in our experiments. \textbf{Step 1} trains the model to predict the next-frame of a given frame sequence. \textbf{Step 2} takes the frozen weights of a model trained in Step 1 (or a randomly initialised model), extracts its latent representations through linear probes, and performs a regression task using the underlying constant \textit{directly}.}
    \label{fig_model_setups}
\end{figure*}

\section{Related Work}
We briefly summarise both the recent history of \textit{video generation} models, and the state of \textit{visual physics modelling} in deep learning.

\subsection{Generative Visual Pretraining}
Though there is well established work in unsupervised feature learning on standalone \textit{images} \cite{Oord2018RepresentationLW, Donahue2019LargeSA}, our methodology is specifically targeted at generating subsequent time-step images as part of \textit{video} sequences. \citet{Ranzato2014VideoM} present an early baseline for such unsupervised video feature learning. Our work is most similar in motivation to that of \citet{Chen2020GenerativePF} who explore unsupervised representation learning for images using a minimal adaption of transformers and vector-quantised Variational Autoencoder (VAE) \citep{Oord2017NeuralDR} architectures. \citet{Chen2020GenerativePF} use auto-regressive next-pixel prediction or masked-pixel prediction training strategies, and present an extensive probing study of the learned representational capacity of layers in their models. They further demonstrate that unsupervised image pretraining leads to state-of-the-art performance on downstream tasks \textit{that increases with model scale}. We however, aim specifically to pair our visual pretraining datasets with \textit{quantifiable} and \textit{qualitatively observable} probing tasks aimed at clearly measuring the physical laws encoded implicitly during next-frame pretraining.


\subsection{Video Generation}
The recurrent CNN proposed by \citet{Ranzato2014VideoM} for unsupervised frame prediction and filling drew inspiration from early language models \citep{Bengio2000ANP} and RNNs \citep{Mikolov2010RecurrentNN}. Model predictions often tend towards still images after a few frames allegedly due to the local spatial and temporal stationary assumption made by the model. The authors note that predicting beyond a few frames inevitably invokes the curse of dimensionality and argue it could necessitate moving from pixel-wise prediction to higher-level pixel cluster features. \citet{Dosovitskiy2016GeneratingIW} propose a family of deep perceptual similarity metrics `DeepSiM' to avoid the `over-smoothed' results of pixel-wise predictions by instead computing distances between image feature vectors. \citet {Amersfoort2017TransformationBasedMO} propose a convolutional network to generate future frames by predicting a transformation based on previous frames and constructing the future frames accordingly, leading to sharper images and simultaneously avoiding the curse of high dimension predictions. \citet{9157218} propose an integrated Bayesian framework to cope with uncertainties caused by noisy observations (\emph{i.e.} perceptual) and forward modelling process (\emph{i.e.} dynamics). \citet{Yilmaz2021DFPNDF} use deformable convolutions \citep{Dai2017DeformableCN} to try and exploit a larger and more adaptive receptive field as opposed to normal convolutions. The recently released VideoGPT \citep{yan2021videogpt} is a video generation model which combines vector-quantised VAE and transformer designs, yielding very high quality frame predictions on the UCF-101 \citep{Soomro2012UCF101AD} and TGIF \citep{Li2016TGIFAN} datasets . However, due to the inherent difficulty of modelling complex real-world long-term videos, errors in motion still build up. We acknowledge the difficulties that even richly resourced models encounter and echo \citet{yan2021videogpt}: videos are just simply a ``hard modelling challenge''. See the work of \citet{Oprea2020ARO} and \citet{9063513} for a thorough review of video generation.

\subsection{Visual Physics Modelling}
\citet{Wu2016Physics1L} collect the `Physics 101' dataset which facilitates models explicitly learning physical properties of objects in videos (\emph{e.g.} mass, acceleration, and friction). Where they focus on encoding physical laws into neural networks, we additionally explore if generative visual modelling is a sufficient or desirable method to induce a quantifiable understanding of these laws. Neural networks have successfully modelled a variety of dynamic systems using images: \emph{e.g.} fluid flow \citep{10.5555/3305890.3306035}, Lyapunov functions \citep{lyapnov}, motion flow \citep{de2018deep} and precipitation nowcasting \citep{articleprecip}. \citet{li2021fourier} propose a novel Fourier neural operator that can learn Burger's equations, Darcy flow, and Navier-Stokes with differing input image resolutions. More recently, \citet{Wang2021MetaLearningDF} push for more generalisable physical modelling with their proposed multi-task DyAd approach.

\section{Models and Configurations}  
\label{models}
We carry out our experiments on two distinct model architectures: a fully convolutional \textit{`CNN'} model with skip connections as a baseline model from vision (Figure \ref{fig_winterbottom_CNN}); and the SegFormer \citep{xie2021segformer} transformer adapted from semantic segmentation to video generation which uses image patches as embeddings and requires multiple patches in sequence to encode the image. We call this adaptation the \textit{`patch transformer'} (Figure \ref{fig:patch_transformer}). Both of these models has been designed or adapted to take a sequence of video frames as input, and output predictions for the next-frame (further described in Section \ref{sec_experiments}).

\subsection{Fully Convolutional 2D CNN}\label{CNN_definition}
\begin{figure*}[!h]
	\centering
	\includegraphics[width=0.7\textwidth]{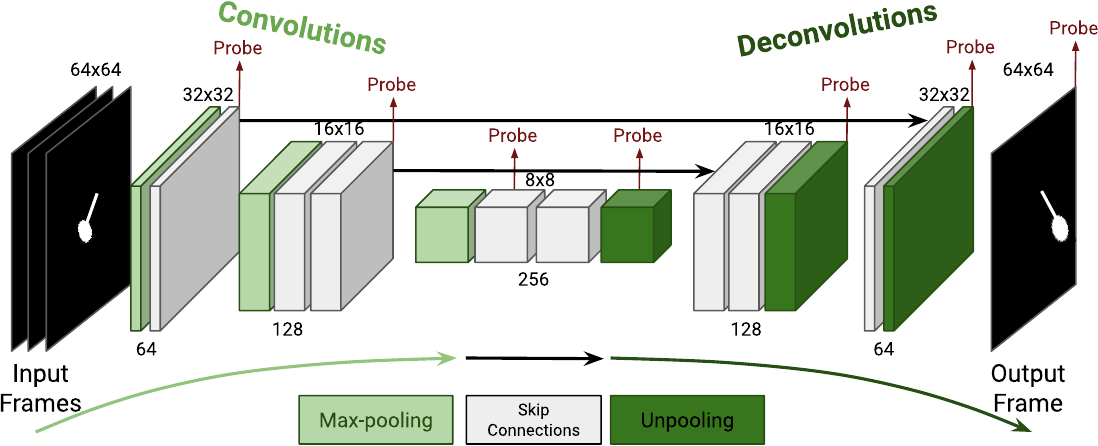}
	\caption{Fully Convolutional 2D \textbf{CNN} model. Each convolution unit is made from two convolution layers, \emph{i.e.} an initial convolution layer that changes the input resolution, followed by another of kernel size 1$\times$1 that does not. The arrows labelled `Probe' indicate which points in the network are extracted to form linear probes used in Section \ref{subsec_probe}.}
	\label{fig_winterbottom_CNN}
\end{figure*}

Introduced by \cite{Long2015FullyCN}, fully convolutional neural networks (FCN) use `deconvolution' layers \citep{Zeiler2014VisualizingAU} \emph{i.e.} convolutional layers with fractional strides. Deconvolution layers can be used to `reverse' the convolution layers and generate a full sized output. Note that upsampling between layers can be learned or can be fixed (\emph{e.g.} bilinear upsampling).
Our CNN baseline model (depicted in Figure \ref{fig_winterbottom_CNN}) is a U-Net style \citep{Ronneberger2015UNetCN} FCN with skip connections \citep{He2016DeepRL}. Given a sequence of $m$ frames of a video, the model takes as input the $m$ 64$\times$64 grayscale input frames in temporal order as inputs for $m$ channels into the first of a series of U-Net style double-convolution units. Each subsequent step halves the image resolution and doubles the number of channels from a starting factor of 64. The upwards pass uses bilinear upsampling and halves the number of channels, mirroring the downward pass in reverse, resulting in the final output frame.

\subsection{Patch Transformer}
\begin{figure*}[!h]
    \centering
    \includegraphics[width=0.9\textwidth]{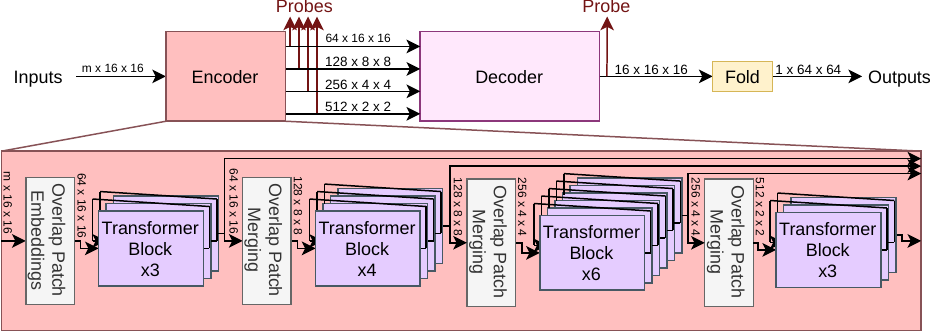}
    \caption{The \textbf{Patch Transformer} model, adapted from the SegFormer \citep{xie2021segformer} for video generation. The arrows labelled `Probe' indicate which points in the network are extracted to form linear probes used in Section \ref{subsec_probe}.}
    \label{fig:patch_transformer}
\end{figure*}

We use a transformer-based semantic segmentation model and modify it minimally to be appropriate for image generation  (Figure \ref{fig:patch_transformer}). Instead of feeding the model RGB images with three channels, we use the sequence of video frames as the input channels and predict the following frame as the output. The patch transformer applies the attention layers across patches of the images. We base the patch transformer architecture on the SegFormer model \citep{xie2021segformer}.

\section{Datasets}
\label{sec_datasets}
As we aim to demonstrate as unambiguously as possible if physical laws are truly induced in predictive training, we create a set of simple dynamic simulations summarised in Table \ref{tab_dsets}.

\begin{table*}[h]
    \centering
        \scalebox{1.0}{
        \begin{tabular}{2c2m{0.27\textwidth}2m{0.56\textwidth}2}
            \Xhline{3\arrayrulewidth}
            \rowcolor{MediumGray}\textbf{Dataset} & \textbf{Example} & \textbf{Affiliated Probing Task(s)}\\
            \Xhline{3\arrayrulewidth}
            \makecell{\textbf{2D Bouncing}\\\#Vids=20,000\\\#Frames=60} & \includegraphics[width=0.28\textwidth]{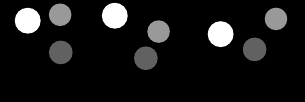} & \textbf{Bounce Regression:} \textit{59 input frames.} Count total bounces demonstrated in the input video. Both ball-to-ball and ball-to-wall bounces, capped at a maximum of 50. \textbf{Gravity Regression:} \textit{5 input frames.} Predict the gravity demonstrated in the given 5 frames. Gravity is in the $y$ axis with 7 potential values [-3e-4, -2e-4, $\cdots$, 3e-4].\\
            \Xhline{3\arrayrulewidth}
            \makecell{\textbf{3D Bouncing}\\\#Vids=10,000\\\#Frames=100} & \includegraphics[width=0.28\textwidth]{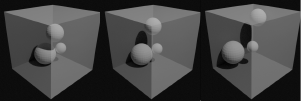} &\textbf{Bounce Regression:} \textit{99 input frames.} Count total bounces demonstrated in the input video. Both ball-to-ball and ball-to-wall bounces, capped at a maximum of 50.\\
            \Xhline{3\arrayrulewidth}
            \makecell{\textbf{Roller}\\\#Vids=10,000\\\#Frames=100} & \includegraphics[width=0.28\textwidth]{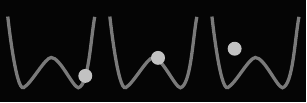} & \textbf{Gravity Regression:} \textit{5 input frames.} Predict the gravity demonstrated in the given 5 frames. Gravity is in the $y$ axis with 201 potential values [0, 0.5, 1, $\cdots$,100].\\
            \Xhline{3\arrayrulewidth}
            \makecell{\textbf{Pendulum}\\\#Vids=10,000\\\#Frames=100} & \includegraphics[width=0.28\textwidth]{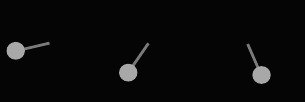} & \textbf{Gravity Regression:} \textit{5 input frames.} Predict the gravity demonstrated in the given 5 frames. Gravity is in the $y$ axis with 41 potential values [0, 0.5, 1.0, $\cdots$, 20].\\
            \Xhline{3\arrayrulewidth}
            \makecell{\textbf{Blocks}\\\#Vids=10,000\\\#Frames=100} & \includegraphics[width=0.28\textwidth]{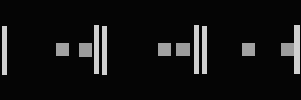} & \textbf{Block Mass Difference Regression:} \textit{49 input frames.} Two blocks of different masses move towards each other on a smooth surface and collide. Predict the difference of the masses between the blocks, positive or negative (positive direction is fixed). Block 1 is always of mass 10, block 2 takes  39 different masses [0.5, 1.0, $\cdots$, 19.5].\\
            \Xhline{3\arrayrulewidth}
            \makecell{\textbf{Moon}\\\#Vids=10,000\\\#Frames=100} & \includegraphics[width=0.28\textwidth]{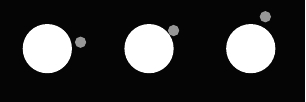} & \textbf{Moon Mass Regression:} \textit{5 input frames.} Predict the gravity demonstrated in the given 5 frames. Gravity acting on the small moon towards the centre of a planet. Masses have 26 potential values [70, 75, 80, $\cdots$,195].\\
            \Xhline{3\arrayrulewidth}
        \end{tabular}
        }
        \caption{Further details of the datasets and their affiliated probing tasks. The constants for estimations for all probing tasks are normalised such that the standard deviation of constants across each individual dataset is 1.}
        \label{tab_dsets}        
\end{table*}

\textbf{2D and 3D Bouncing Balls:} The 2D bouncing dataset consists of videos with 1–3 balls that can collide with both each other and the borders of the image. In this explicit Euler simulation, we vary: the number of balls, ball radius (per ball), initial position (per ball), initial velocity (per ball), gravitational strength, gravity direction, background colour, and ball colour (per ball). We define two probing tasks on this 2D dataset: \textit{total number of bounces estimation} and \textit{$y$-directional gravity estimation}. We also extend this approach to 3D, rendering the balls using realistic lighting from a single light source. This 3D scenario is designed to be a more challenging dataset as balls occlude each other and cast shadows on both the environment boundary and on other balls. We define one probing task for this 3D bouncing dataset: \textit{total number of bounces estimation}.

We experiment with the following simulations based on the myphysicslab project \footnote{https://www.myphysicslab.com}. These simulations are calculated using the Runge-Kutta method \citep{butcher1996history}:

\textbf{Mars Moon:} A simplified simulation of an asteroid orbiting a moon using a rigid body simulation. We vary the initial velocity, moon radius, moon mass, and asteroid radius. The probing task is to \textit{predict the mass of the moon}.

\textbf{Colliding Blocks:} Simulates two blocks that move along a single axis colliding with both the boundary walls and each other. We vary the masses of \textit{one} of the blocks (leaving the other block mass fixed), the starting positions, and starting velocities. The probing task is to \textit{predict the difference in the masses of the two blocks}.

\textbf{Pendulum:} A single pendulum modelled as a point mass at the end of a massless rod. We vary the initial angle, gravity strength, pendulum length, and pendulum mass. The probing task is to \textit{predict the gravity acting on the pendulum}.

\textbf{Roller Coaster with Flight:} A ball of mass $M$ is released down a curved track under gravity $g$ following $F = M\cdot g\cdot \cos({\theta})$. The ball can switch to free flight when the acceleration normal to the curve is greater than $v^2/k$ (where $v$ is the velocity of the ball, and $k$ is the radius of curvature at the current point along the curve). We vary the gravity strength and track position. The probing task is to \textit{predict the strength of gravity acting on the ball}.

\section{Experiments}
\label{sec_experiments}

\subsection{Generative Frame Prediction} 
\label{sec_gen_fram_pred}
Given a sequence of $m+1$ frames of a video, the model takes as input the first $m$ 64$\times$64 grayscale input frames in temporal order and predicts as output the next-frame. This predicted frame is compared against the final (ground truth) frame in the $m+1$ sequence. These experiments are visualised as \textbf{Step 1} in Figure \ref{fig_model_setups}.
We use either smooth-$L$1 (SL1) or Structural Similarity (SSIM) as loss functions. As SSIM takes values between -1 and 1, and should be maximised, we reformulate it as a minimisation problem (as in Equation \ref{equation_ssim_loss}) in order to use it as a loss function: 
\begin{equation}
\label{equation_ssim_loss}
    \mathcal{L}_{\text{SSIM}} = 1-\frac{1+SSIM}{2}  \in [0,1].
\end{equation}
Using the models introduced in Section \ref{models}, we perform modelling experiments on each of the 6 datasets introduced in Section \ref{sec_datasets} using both SSIM and SL1 loss. The values of $m$ in our experiments (\emph{i.e.} number of input frames) depends on the probing task it will be paired with. See Table \ref{tab_dsets} for further details. To assess the of performance the modelling task, we consider the Peak Signal-to-Noise Ratio (PSNR) \citep{5596999}, Structural Similarity (SSIM) \citep{1284395}, and $L$1 scores between the predicted frame and the ground truth frame. We do not consider metrics such as Learned Perceptual Image Patch Similarity (LPIPS) \citep{8578166}  and Fr\'{e}chet Video Distance (FVD) \citep{unterthiner2019fvd}. LPIPS is a deep model based metric that is only well defined on RGB images, and FVD requires an image resolution of at least 224$\times$224.

\subsection{Physical Constant Estimation Probing Tasks}
After the generative frame prediction task, we wish to assess the extent to which features capable of predicting the underlying physical constants have been indirectly induced. To achieve this, we extract the frozen features at multiple points in each model, concatenate them together, and use them as inputs for the physical constant estimation regression task (see \textbf{Step 2} in Figure \ref{fig_model_setups}). In order to quantify how well the physical constant estimation tasks perform, we must have some notion of what a poor or `random' performance is for the task. However, since these physical constant estimation tasks are regression of single arbitrary values (whatever arbitrary units of gravity/mass we have chosen), it is initially impossible to tell if a loss of 0.2, or 0.002 can be thought of as either good or random performance. We address this ambiguity with the following: We normalise the values for the ground truths for all 6 datasets such that each has a standard deviation of 1 across its labels; We provide a notion of poor or `random' performance on each of these tasks through 2 \textbf{lower-bound} baselines (see Section \ref{sec_exp_baselines}). We apply a smooth-$L$1 loss with $\beta = 0.01$ for all tasks.

\subsubsection{Lower-Bound Baselines} 
\label{sec_exp_baselines}
\begin{figure*}[t]
    \centering
    \includegraphics[width=1.0\textwidth]{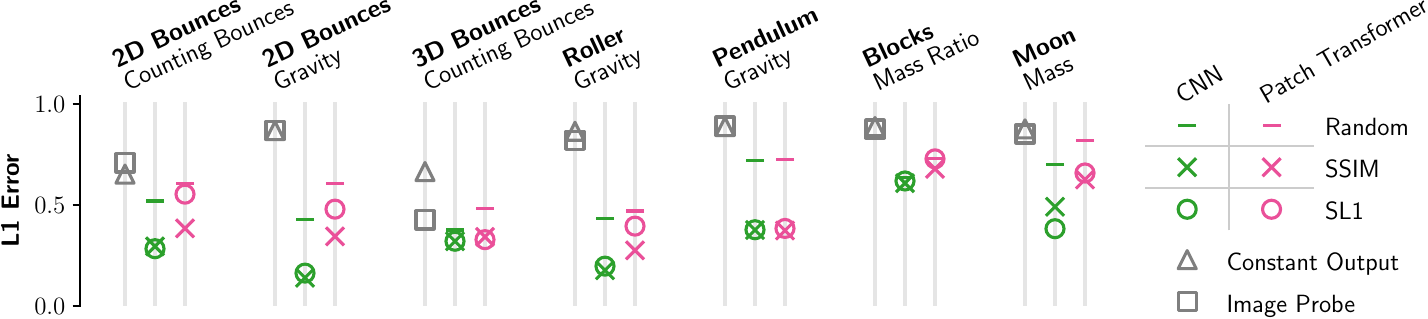}
    \caption{The results of both our lower bound and linear probing probing task experiments. The two lower bound setups indicated the grey squares and triangles are described in Section \ref{sec_exp_baselines}. The coloured cross and circle indicate the probed model has been pretrained on SSIM and SL1 loss respectively.} 
    \label{fig_probing_results}
\end{figure*}

\textbf{Optimal Constant Output:} We calculate what the best possible loss would be if the model were forced to predict a constant (but optimally chosen) value, \textit{i.e.} an `average' value of the ground truths across the dataset (grey triangles in Figure \ref{fig_probing_results}).

\textbf{Image+Linear Layer:} We flatten the input images and pass them into a simple trainable linear layer, the output of which we use to calculate the probing task loss (grey squares in Figure \ref{fig_probing_results}).

\subsubsection{Probing Frozen Models}
\label{subsec_probe}
We freeze the weights of a model that has been pretrained on the generative frame prediction task (described in Section \ref{sec_gen_fram_pred} and \textbf{Step 1} on Figure \ref{fig_model_setups}) and flatten and concatenate the outputs of each substantial layer throughout the network (see `Probe' arrows in Figures \ref{fig_winterbottom_CNN} and \ref{fig:patch_transformer}) and send these features through a single trainable linear layer (as to extract only the information already learned), before applying SL1 loss to the output.


\section{Results and Discussion}

In this section, we discuss the results of our generative pretraining and probing task experiments.
\subsection{Generative Frame Prediction Performance}
\begin{figure}[!t]
    \centering
     \includegraphics[width=0.45\textwidth]{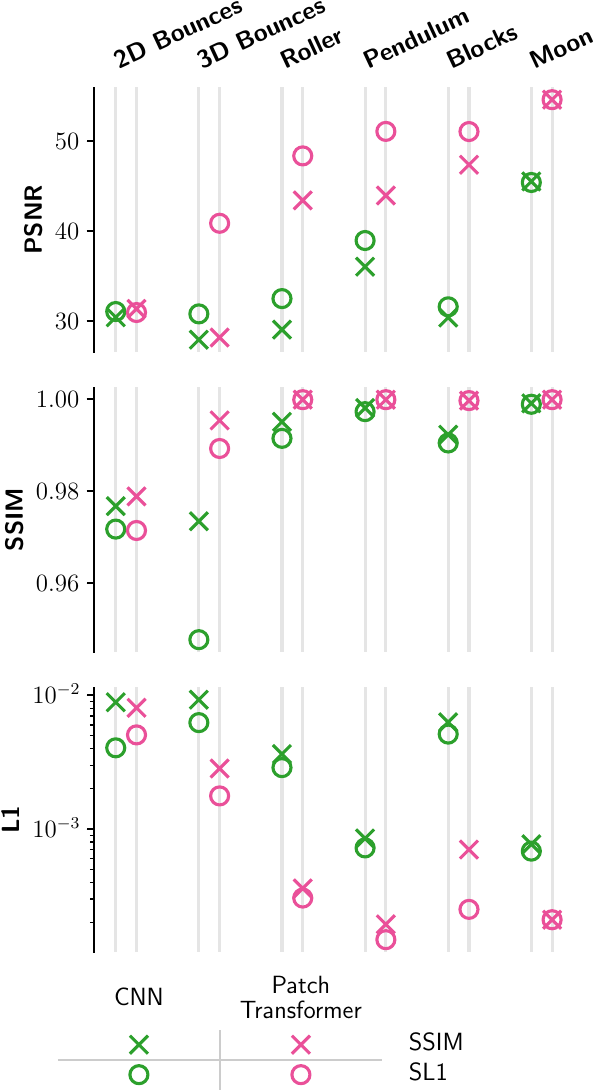}
    \caption{Metrics calculated between the ground truth and the predicted frame on each modelling dataset. Higher is better for PSNR and SSIM, and lower is better for \textit{L}1.} 
    \label{fig_metrics}
\end{figure}

We consider the quality of the generated frames by computing standard image-generation metrics ---PSNR, SSIM and $L$1 scores--- between the predicted frame and the ground truth frame (Figure \ref{fig_metrics}). We compare these scores with respect to the different datasets, models, and losses. Such quantitative metrics are important standards to measure the quality of image generation in machine learning. However, as we highlight here, these metrics have their limitations and when viewed in isolation cannot always reflect the nuanced behaviours of dynamic simulations. We therefore augment these \textit{quantitative} measures with \textit{qualitative} analyses of our model's understanding of our dynamic simulation datasets. We do this iterating forcing our models a further 19 frames into the future, giving a total of 20 frames of prediction with which to visually compare to the ground truth of the rest of the simulation, discussed in Section \ref{res_selfout}.

\textbf{Generative Prediction Quality - Dataset:}
We see that the modelling tasks yielding the best scores are roller, pendulum, blocks, and moon. We believe this is because the background for these tasks is always black, and there are fewer structural variations than the other datasets. A consistent background colour and structure ---which comprise the majority of the overall image--- allows the model to very accurately predict the majority of the pixels near perfectly, leading to better overall scores in the metrics. The 2D and 3D bouncing datasets yield scores slightly lower than the top performing four datasets. Though we expected the 3D bouncing dataset to be more complicated than the 2D version given the extra dimension of movement to understand, we actually find that the 3D bouncing dataset yields slightly better scores than the 2D bouncing dataset. We find that although 3D bouncing must model motion in an extra dimension, the 2D bouncing dataset has deceptively challenging elemnts: varying background and ball colours; varying gravitational constants; and an increased number of interactions between balls and walls.

\textbf{Generative Prediction Quality - Model:} The patch transformer consistently has the best PSNR, SSIM, and $L$1 metric scores. Notably, the patch transformer's lead over the CNN widens considerably on the four datasets with best scores overall (roller, pendulum, blocks, and moon).

\textbf{Generative Prediction Quality - Loss:} As we would intuitively expect, we see that models trained with the SSIM-based \textit{loss} also demonstrate higher SSIM \textit{scores} on their predictions compared to those pretrained on the SL1 loss. Conversely, models trained with the mean pixelwise SL1 loss have a lower (\emph{i.e.} better) mean pixelwise $L$1 score. This improvement in each metric of models trained with that metric's respective loss counterpart further highlights how limited any single metric is in demonstrating the quality of generated images. Despite each score's `preference' for its own loss function, we find that SL1 trained models almost always give a higher PSNR than their SSIM counterparts. This more pronounced covariance of PSNR and SL1 scores is expected behaviour as PSNR tends to infinity as mean squared error (and hence SL1) tends to zero, and thus we expect that PSNR is optimised for by an SL1 loss.

\begin{figure*}
    \centering
    \begin{subfigure}[b]{\textwidth}
        \centering
        \includegraphics[width=0.9\textwidth]{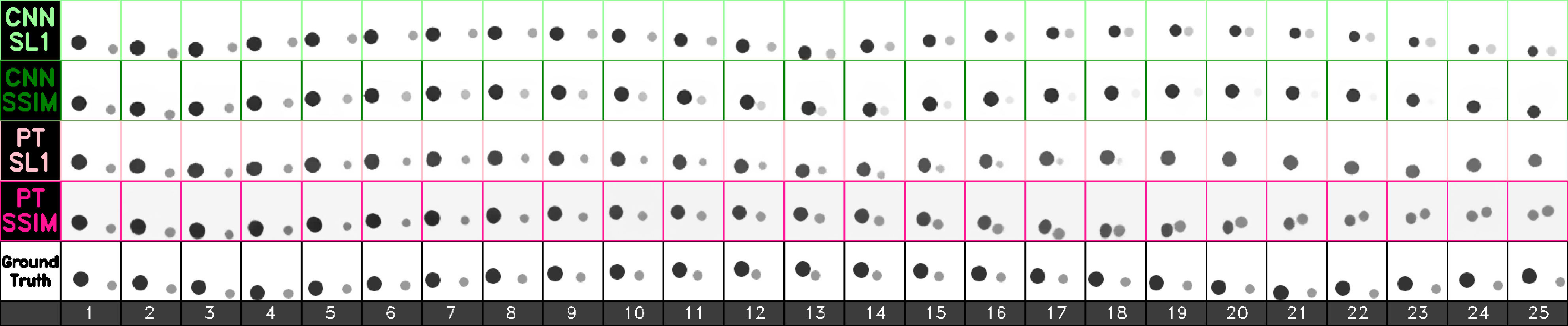}
        \caption{2D Bouncing}
        \label{fig_so_2db}
    \end{subfigure}
    \hfill
    \begin{subfigure}[b]{\textwidth}
        \centering
        \includegraphics[width=0.9\textwidth]{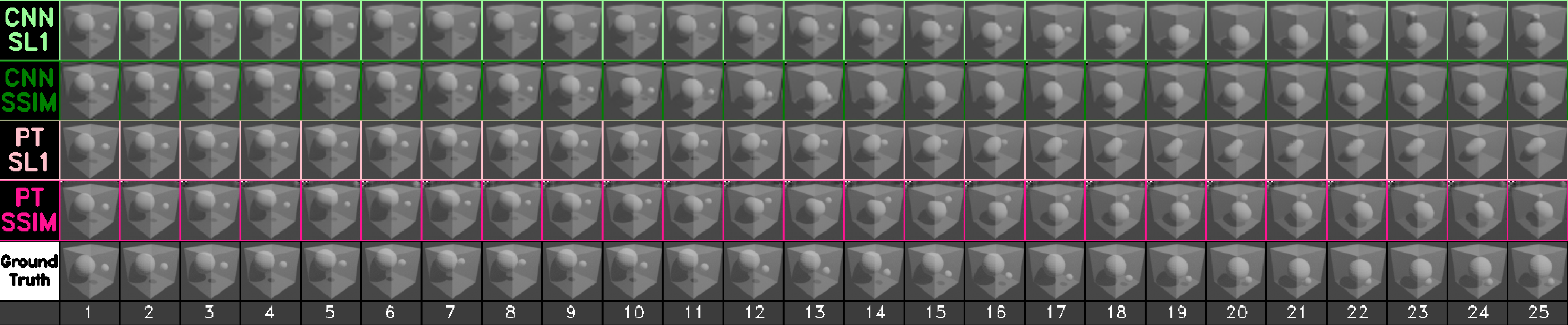}
        \caption{3D Bouncing}
        \label{fig_so_3db}
    \end{subfigure}
    \hfill
    \begin{subfigure}[b]{\textwidth}
        \centering
        \includegraphics[width=0.9\textwidth]{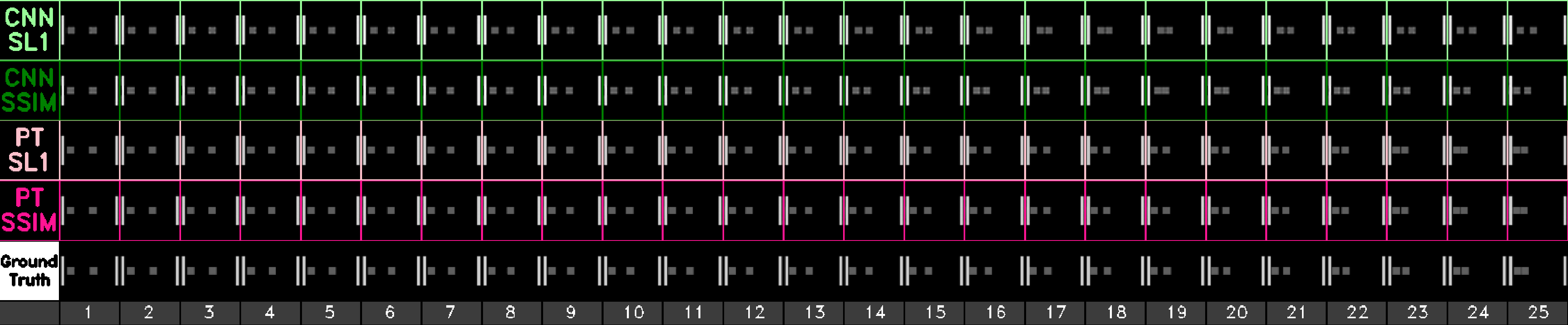}
        \caption{Blocks}
        \label{fig_so_blocks}
    \end{subfigure}
    \hfill
    \begin{subfigure}[b]{\textwidth}
        \centering
        \includegraphics[width=0.9\textwidth]{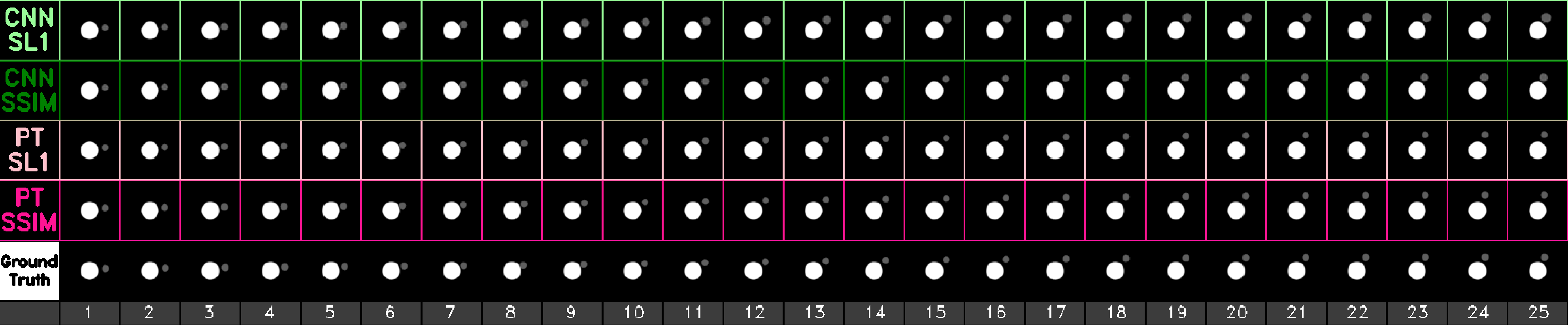}
        \caption{Moon}
        \label{fig_so_moon}
    \end{subfigure}
    \hfill
    \begin{subfigure}[b]{\textwidth}
        \centering
        \includegraphics[width=0.9\textwidth]{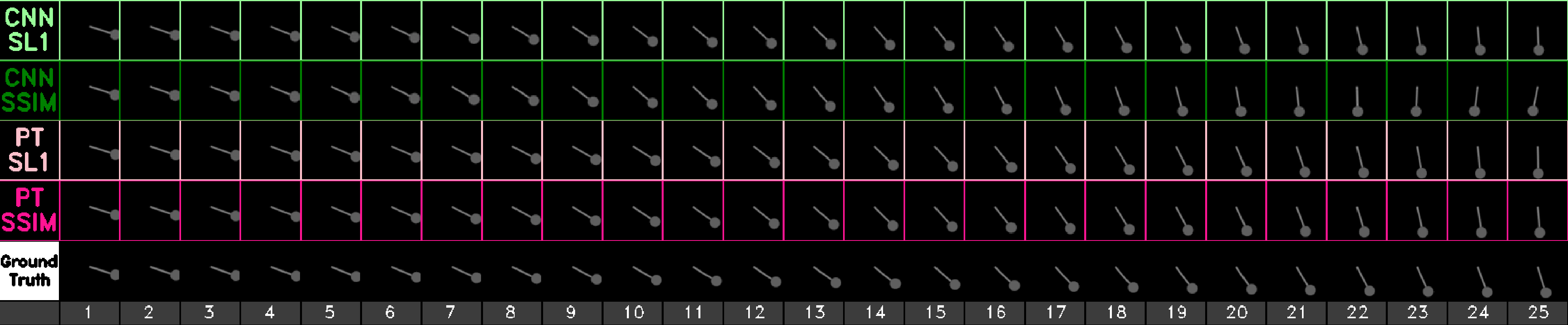}
        \caption{Pendulum}
        \label{fig_so_pendulum}
    \end{subfigure}
    \hfill
    \begin{subfigure}[b]{\textwidth}
        \centering
        \includegraphics[width=0.9\textwidth]{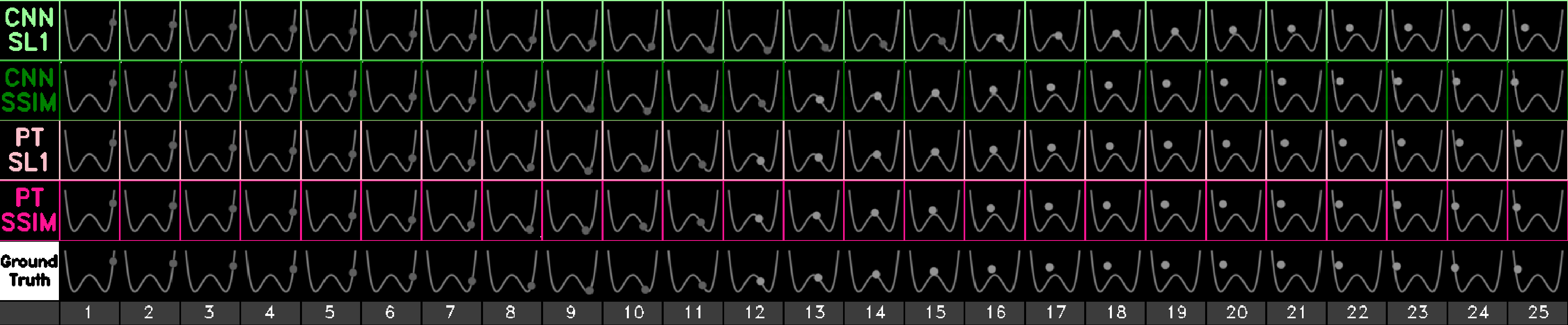}
        \caption{Roller}
        \label{fig_so_roller}
    \end{subfigure}
    \vspace{-2em}
    \caption{Comparison of the first 25 generated frames of each model ($m=5$) visualised alongside the ground truth.} 
    \vspace{-1em}
    \label{fig_so}
\end{figure*}

\subsection{Long-Term Prediction}
\label{res_selfout}
We complement these quantitative measures of frame prediction with a qualitative analysis of the `long-term' prediction capabilities by generating an extra 20 frames into the future. At each such step, the model's previous predicted frame becomes the new final frame of the inputs and the first frame of the original inputs are discarded. By comparing this generated video sequence with the ground truth (\textit{e.g.} Figures \ref{fig_so_2db} and \ref{fig_so_3db}), we can qualitatively assess how well the model has understood the physical laws underpinning the 6 datasets. We invite readers to explore our complete set of self-output videos and metrics for each test set \footnote{\url{https://github.com/Visual-modelling/visual\_modelling\#all-self-output-gifs}}.

\textbf{Long-Term Prediction - Models:} The patch transformer generates the most accurate long-term predictions on the 3D bouncing, roller, pendulum, blocks, and moon datasets. The CNN model performs best on the 2D bouncing (Figure \ref{fig_so_2db}) dataset. The CNN struggles to smooth out lower resolution graininess. There are occasional oddities with the CNN that imply it is overly relying on local spatial information, \emph{e.g.} a ball will sometimes accelerate off the screen in the roller dataset. This could be because the model has learned that a ball above a rail at a certain angle should be moving upwards. Although it is thought that pure CNN architectures struggle to properly model inter-frame variations in video sequences \citep{Oprea2020ARO}, our CNN model's self-output videos challenge this assumption by demonstrating an understanding of gravity and collision physics (Figure \ref{fig_so_2db}). Though the patch transformer appears to be the best model, it too demonstrates architecture specific artifacts (\emph{e.g.} the resolution of the patches forming the outputs are visible in some examples).

\textbf{Long-Term Prediction - Loss:} 
We find that training on SSIM loss causes both our models to distort the otherwise constant background colour (seen in the SSIM rows of Figure \ref{fig_so_2db}). This phenomenon implies a gradual build-up of errors in the background which we argue is due to the insensitivity of SSIM function to changes in the background, and thus the model is not forced to carefully maintain the background. Such background distortion does not happen in SL1 outputs, indicating that the pixelwise-SL1 loss is particularly sensitive to these artefacts and prioritises minimising them.

\subsection{Probing Task Performance}

Both the CNN and patch transformer probing tasks on pretrained models consistently score better than our lower bounds. This implies that generative pretraining has encoded some understanding that is useful to our probing probing tasks. There are few noticeable differences between SL1 and SSIM pretraining for the CNN. SSIM is marginally better on 2D bouncing gravity estimation (by 0.03), and SL1 is 0.11 better for the moon task. The patch transformer however demonstrates improved performance for SSIM pretraining on all but the 3D bouncing bounce counting task, improving by between (0.03-0.15).

\section{Limitations and Future Directions}
In this section, we discuss the main limitations of our study and highlight areas which we believe are promising for future research.

\textbf{Scale:} Our generative pretraining approach does not match the sheer scale of modern language modelling. Modern language models train on a huge amount of data due to the comparative ease of collecting, storing, and tokenising raw text. Such language models are often trained with substantial computational resources. By directly training on and outputting raw images, which are much larger representations than language tokens, we can't yet approach the scale of large pretrained modern language models with the resources available to us.

\textbf{Dataset Complexity:} Larger and more `visually complex' video datasets are unexplored. We have prioritised visually simple scenarios to make our analyses as quantitatively and qualitatively verifiable as possible. Nonetheless, a natural next step will be to raise the visual complexity and expand the scope of the `physical laws' to be demonstrated. To the best of our knowledge, there are no physical dynamics dataset more `visually complex' than ours that are paired with quantified measures of their underlying physical laws. This would be the next kind of dataset to design and create.   


\textbf{More Nuanced Predictive Training Strategy:} Though we parallel language modelling, we only predict the final frame of a sequence, and do not parallel more intricate language modelling training strategies, \emph{e.g.} bidirectional token prediction for words in the middle of a sentence. Although more nuanced training methods for video has been explored on techniques from several years ago \citep{Ranzato2014VideoM}, and more recently for generative pretraining with \textit{images} \citep{Chen2020GenerativePF}, we are able to present strong pretraining results without using such nuanced approaches in our \textit{video-based} study. Regardless, this is an interesting line of research for future work.

\section{Conclusion}
We explore generative frame prediction as a mechanism for efficiently inducing a \textit{verifiable} understanding of `physical laws'. We introduce 6 dynamic simulation video datasets that also function as physical property estimation datasets with a total of 7 probing tasks. The \textit{pairing} of our visual pretraining datasets with \textit{quantifiable} and \textit{observable} probing tasks let us demonstrate the inductive potential of generative pretraining. Our generative frame prediction experiments demonstrate an understanding of the laws underpinning all 6 datasets has been achieved by both models. Our linear probing experiments further demonstrate that the frame prediction task induces a \textit{quantifiable} understanding of these physical laws through a significant improvement over reasonable baselines in the paired probing tasks. Further improving the scale of visual modelling pretraining is a promising path to greater visual understanding.

\section*{Acknowledgments}
We would like to thanks William Prew and James Bower for their helpful insights.

\bibliographystyle{plainnat}
\bibliography{ref}

\end{document}